%% file: example_paper.tex
\documentclass{article}

\usepackage{microtype}
\usepackage{graphicx}
\usepackage{booktabs} 

\input{math_commands.tex}   %
\usepackage{mathtools}      
\usepackage{amsthm}         
\usepackage{nicefrac}       

\usepackage{thmtools,thm-restate}

\usepackage{siunitx} 
\usepackage{multirow}
\usepackage{array}
\usepackage{caption}
\usepackage{subcaption}
\usepackage{graphicx}
\usepackage{bbm} 

\usepackage{xspace}
\newcommand*{\eg}{\textit{e.g.}\@\xspace}

\newcommand*{\ie}{\textit{i.e.}\@\xspace}

\newcommand*{\wrt}{w.r.t.\@\xspace}

\usepackage{hyperref}

\usepackage[accepted]{icml2021}

\icmltitlerunning{Consistency Regularization Can Improve Robustness to Label Noise}

\begin{document}

\twocolumn[
\icmltitle{Consistency Regularization Can Improve Robustness to Label Noise}



\icmlsetsymbol{equal}{*}

\begin{icmlauthorlist}
\icmlauthor{Erik Englesson}{ed}
\icmlauthor{Hossein Azizpour}{ed}
\end{icmlauthorlist}

\icmlaffiliation{ed}{Division of Robotics, Perception, and Learning, KTH, Stockholm, Sweden}

\icmlcorrespondingauthor{Erik Englesson}{engless@kth.se}

\icmlkeywords{Machine Learning, ICML}

\vskip 0.3in
]



\printAffiliationsAndNotice{}  

\input{0_abstract}
\input{1_new_introduction}

\input{3_method}

\input{4_results}
\input{2_related_works_short}

\input{5_future_work_limitations}
\input{acknowledgement}
\bibliography{references}

\input{7_appendix}

\bibliographystyle{icml2021}

%
%
%

\end{document}

%% file: math_commands.tex
\usepackage{amsmath,amsfonts,bm}

\def\eqref#1{equation~\ref{#1}}

\def\1{\bm{1}}

\def\vm{{\bm{m}}}

\def\vp{{\bm{p}}}

\def\vx{{\bm{x}}}
\def\vy{{\bm{y}}}

\DeclareMathAlphabet{\mathsfit}{\encodingdefault}{\sfdefault}{m}{sl}
\SetMathAlphabet{\mathsfit}{bold}{\encodingdefault}{\sfdefault}{bx}{n}

\def\sX{{\mathbb{X}}}
\def\sY{{\mathbb{Y}}}

\DeclareMathOperator*{\argmax}{arg\,max}

\newcommand{\DJS}{\mathrm{JS}_{\pi}}

\newcommand{\DJSnp}{\mathrm{JS}}
\newcommand{\DGJSnp}{\mathrm{GJS}}


%% file: 0_abstract.tex
\begin{abstract}
Consistency regularization is a commonly-used technique for semi-supervised and self-supervised learning. It is an auxiliary objective function that encourages the prediction of the network to be similar in the vicinity of the observed training samples. \citet{hendrycks2020augmix} have recently shown such regularization naturally brings test-time robustness to corrupted data and helps with calibration. This paper empirically studies the relevance of consistency regularization for training-time robustness to noisy labels. First, we make two interesting and useful observations regarding the consistency of  networks trained with the standard cross entropy loss on noisy datasets which are: (i) networks trained on noisy data have lower consistency than those trained on clean data, and (ii) the consistency reduces more significantly around noisy-labelled training data points than correctly-labelled ones. Then, we show that a simple loss function that encourages consistency improves the robustness of the models to label noise on both synthetic (CIFAR-10, CIFAR-100) and real-world (WebVision) noise as well as different noise rates and types and achieves state-of-the-art results.
\end{abstract}

%% file: 1_new_introduction.tex
\vspace{-0.4cm}
\section{Introduction}
\label{sec:intro}
Labelled datasets, even the systematically annotated ones, contain  noisy labels~\citep{Beyer_arXiv_2020_done_ImageNet}. One key advantage of deep networks and stochastic-gradient optimization is the resilience to label noise~\citep{Li_AIStats_2020_DeepLearningRobust}. Nevertheless, it has been frequently shown that this robustness can be significantly improved via a noise-robust design of the model~\cite{Vahdat_NeurIPS_2017_CRF,Li_ICLR_2020_dividemix,Seo_NeurIPS_2019,Iscen_ECCV_2020,Nguyen_ICLR_2020_self_ensemble}, the learning algorithm~\cite{Reed_arXiv_2014_bootstrapping,Northcutt_arXiv_2017,Tanaka_CVPR_2018_Joint_Optimization,Lukasik_ICML_2020_label_smoothing_label_noisy,liu2020earlylearning} or the provably-robust loss function~\cite{Ghosh_AAAI_2017_MAE,Wang_ICCV_2019_Symmetric_CE,Zhang_NeurIPS_2018_Generalized_CE,Ma_ICML_2020_Normalized_Loss,Liu_ICML_2020_Peer_Loss}. In this work, we focus on a technique, called \textit{consistency regularization}, for robustness to training label noise.

Consistency regularization is a recently-developed technique that encourages smoothness of the learnt function. It has become increasingly common in the state-of-the-art semi-supervised learning~\citep{Miyato_PAMI_2018_VAT,Berthelot_NeurIPS_2019_mixmatch,Tarvainen_NIPS_2017_mean_teacher} and test-time robustness to input corruptions~\citep{hendrycks2020augmix}. Recently, DivideMix~\citep{Li_ICLR_2020_dividemix} used consistency regularization in an elaborate pipeline for label noise via the use of a semi-supervised method.

\begin{figure*}
        \centering
        \begin{subfigure}[b]{0.45\textwidth} \centering \includegraphics[width=0.7\textwidth]{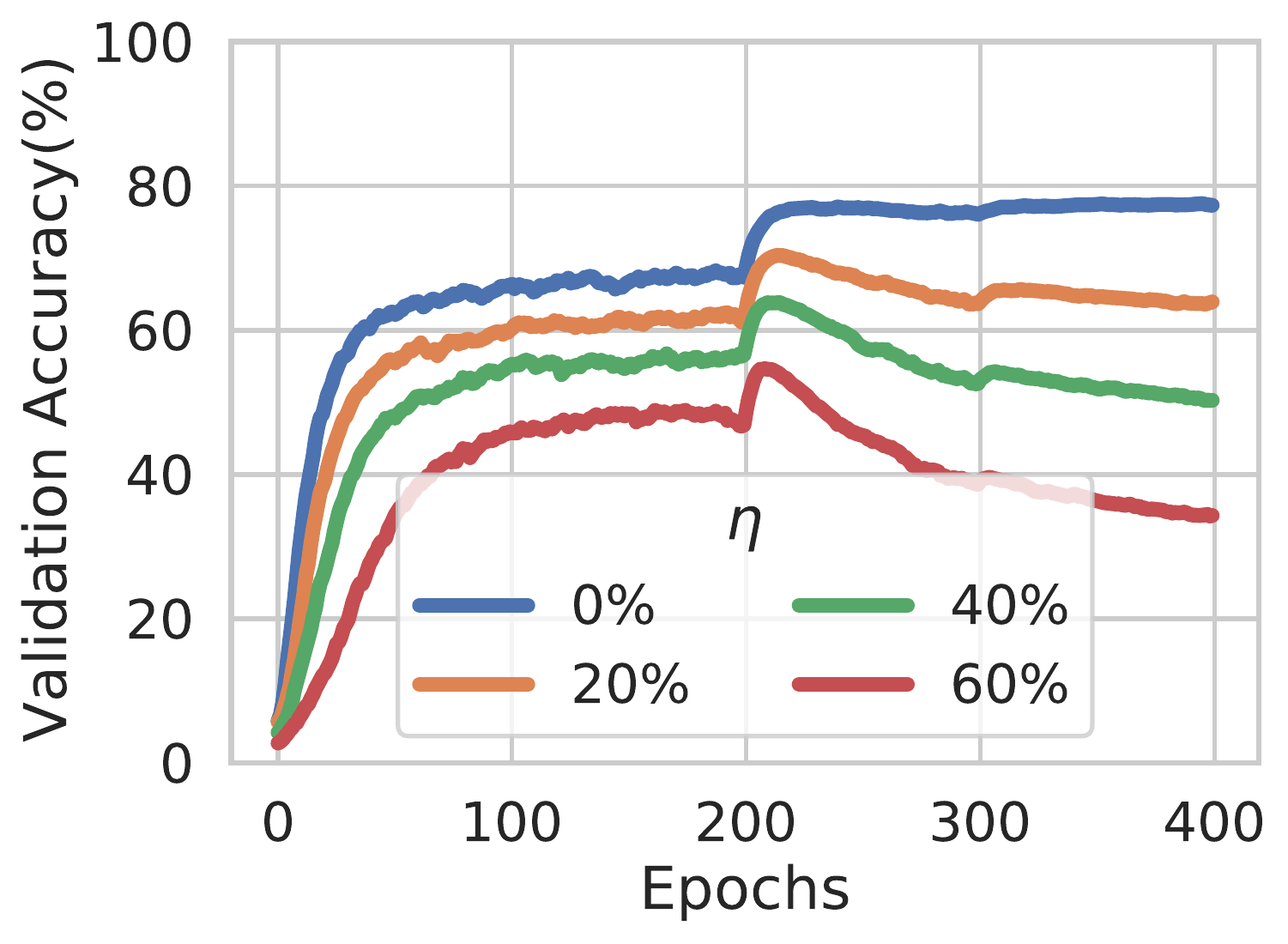} \caption{Validation Accuracy} 
         \end{subfigure} 
         \quad
         \begin{subfigure}[b]{0.45\textwidth} \centering \includegraphics[width=0.7\textwidth]{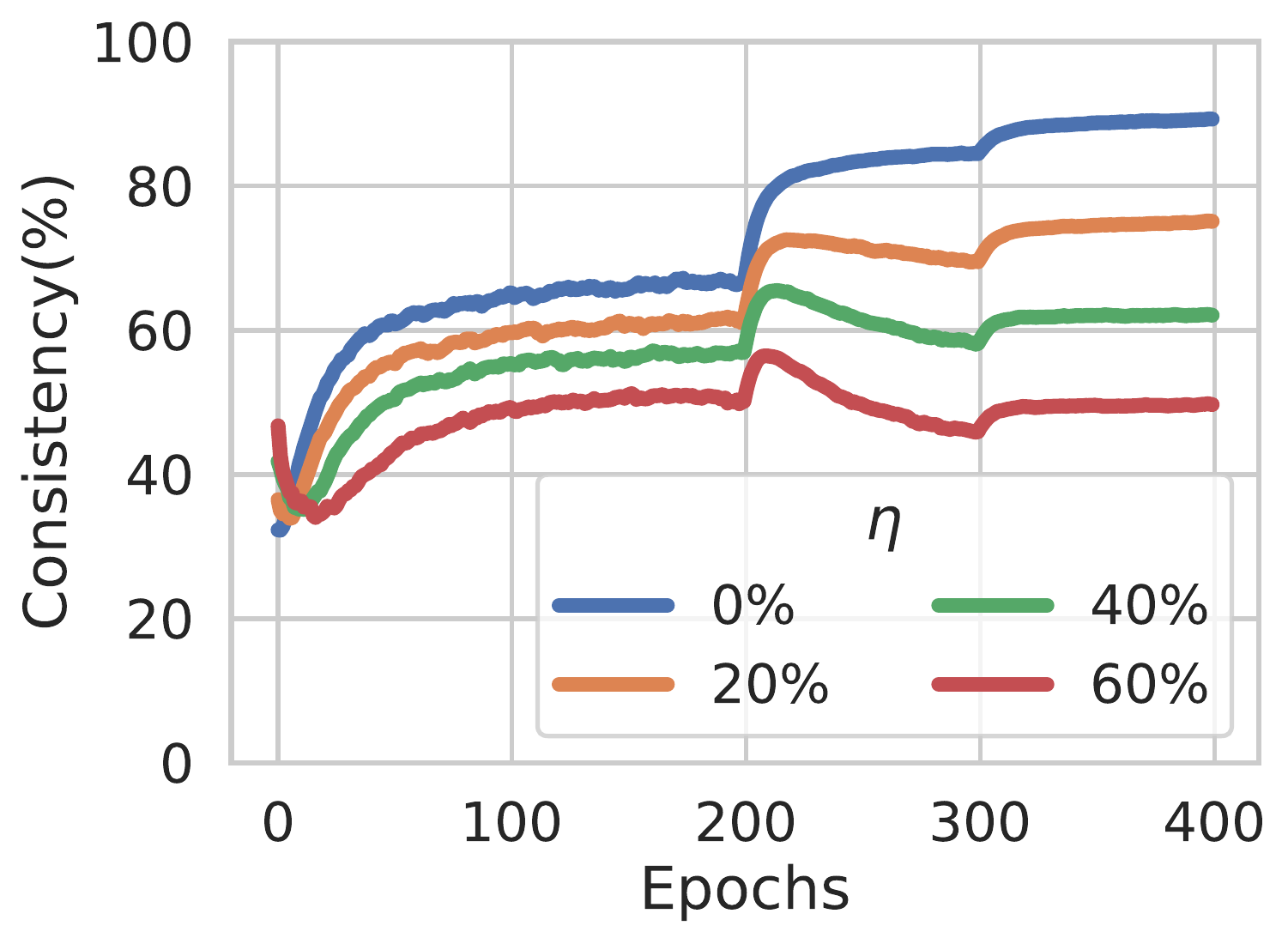} \caption{Consistency} 
         \end{subfigure} 
         \caption{\textbf{Evolution of a trained network's consistency as  it overfits to noise using CE loss.} Here we plot the evolution of the validation accuracy~(a) and network's consistency~(b) 
         on the training set of CIFAR-100 for varying symmetric noise rates when learning with the cross-entropy loss. The consistency of the learnt function and the accuracy closely correlate. Assuming causation, suggests that enforcing consistency may help avoid fitting to noise. 
         } \label{fig:consistency-observation}
         \vspace{-0.2cm}
\end{figure*}

In this work, we solely focus on the relevance of a network's learnt function consistency for robustness to training label noise. First we make two novel  and interesting observations: (i) networks trained on noisy data exhibit a generally lower consistency than those trained on clean data (Figure~\ref{fig:consistency-observation}), and (ii) the consistency reduces more significantly around noisy-labelled training data points than correctly-labelled ones (Figure~\ref{fig:consistency-clean-noisy}). These important observations empirically motivate the use of consistency regularization for robustness to label noise. Thus, we adopt a simple loss function, similar to AugMix~\citep{hendrycks2020augmix}, to improve  training-time robustness to noisy-labelled data. Doing so, we achieve remarkable performance on the synthetically-noisy versions of CIFAR-10 and CIFAR-100 using both symmetric and asymmetric noise at various rates. Furthermore, we show state-of-the-art performance on the real-world noisy dataset of WebVision comparable to that of DivideMix which benefits from a significantly more complicated pipeline.

%% file: 3_method.tex
\section{Consistency for Learning with Label Noise}
\label{sec:GJS}

In this section, we first make two intriguing observations which motivate the use of consistency regularization for learning with noisy labels. Then we propose a loss function to this end preceded with some necessary background.
\subsection{Consistency When Overfitting to Noise}
\label{sec:consistency-obs}
In Figure \ref{fig:consistency-observation}, we show the evolution of (a) the validation accuracy and (b) a measure of consistency during training with the CE loss for varying amounts of noise. First, we note that training with CE loss eventually overfits to the noisy labels. 
Figure~\ref{fig:consistency-observation}(b) shows the consistency of predictions \wrt the CIFAR-100 training set. Consistency is measured as the fraction of images that have the same class prediction for the original image and an augmented version of it, see Appendix~\ref{sup:sec:consistencyMeasure} for more details. Interestingly, a clear correlation is observed between validation accuracy and training consistency. Assuming causality, suggests that maximizing consistency of predictions may improve robustness to noise. 

In Figure~\ref{fig:consistency-clean-noisy}, similarly to Figure~\ref{fig:consistency-observation}, we study the consistency among training examples, but separately for those with noisy/wrong labels and those with clean/correct labels. Crucially, we find that the consistency around noisy examples degrades more significantly than for clean examples. This suggests that encouraging noisy examples to stay consistent will make it harder to overfit to noise. 

Based on these observations, we propose a loss function to encourage consistency in Section~\ref{sec:gjs-def}. Before presenting the loss function, we briefly reiterate some background.

\subsection{Background}
\textbf{Supervised Classification.}  
Given a training dataset $\mathcal{D} = \{ (\vx_i,y_i)\}_{i=1}^N$, our goal is to learn the parameters of a softmax neural network~($f$) mapping each $\vx \in \sX$ to its corresponding onehot representation~($\vy$) of the class $y \in \sY = \{1,2,\dots,K\}$. The function $f$ is trained on $\mathcal{D}$ by minimizing an empirical risk $\frac{1}{N}\sum_{i=1}^N \mathcal{L}(\vy_i,f(\vx_i))$, where $\mathcal{L}$ is a loss function.

\textbf{Learning with Noisy Labels.} In this work, we learn from a noisy training distribution $\mathcal{D}_\eta$ where the labels are changed, with probability $\eta$, from their true distribution $\mathcal{D}$. The noise is called \textit{symmetric} if the noisy label is independent of the true label and \textit{asymmetric} if it is class dependent. The considered noise in this work is sample-independent.

\subsection{Consistency Loss}
\label{sec:gjs-def}
There are many ways to encourage consistency between predictions. Most methods combine a label-dependent term~(typically cross-entropy) with a separate label-independent consistency term based on $l^2$-norm~\citep{Tarvainen_NIPS_2017_mean_teacher}, CE~\citep{Miyato_PAMI_2018_VAT}, etc. Here, similar to AugMix~\cite{hendrycks2020augmix}, we use a loss based on Jensen-Shannon divergence to encourage consistency.

Let $\vy$ be the onehot label, and $\vp_1,\vp_2$ be predictions from two augmentations of the same image, and a weight $\pi$ satisfying $0~\leq~\pi~\leq 1$,  then our loss is
{
\begin{align}
\DGJSnp \coloneqq \mathrm{JS}_{\pi}(\vy,\frac{\vp_1+\vp_2}{2}) 
+ (1-\pi)\mathrm{JS}_{\frac{1}{2}}(\vp_{1},\vp_{2}) \label{eq:gjs-loss}
\end{align}
}%
where $\DJSnp_{\pi}$ is the Jensen-Shannon divergence
{
\begin{align}
    \mathrm{JS}_{\pi}(\vp_{1}, \vp_{2}) \coloneqq 
    \pi\mathrm{KL}(\vp_{1} \Vert \vm) + (1-\pi)\mathrm{KL}(\vp_{2} \Vert \vm) 
    \label{eq:GJSasKL}
\end{align}
}%
with the mean $\vm \coloneqq \pi\vp_1 + (1-\pi)\vp_2$. 

The two $\DJS$ terms of $\DGJSnp$ in  Equation~\ref{eq:gjs-loss} encourage the predictions to be close to the label and to be consistent, respectively. For the label-dependent term, we treat $\pi$ as a hyperparameter, while the consistency term uses equal weights~($\pi=\frac{1}{2}$). We also use the loss $\DJSnp \coloneqq \DJS(\vy, \vp_1)$ as a baseline to compare with a Jensen-Shannon-based loss without consistency. Indeed, $\DGJSnp$ generalizes $\DJSnp$ by incorporating consistency regularization. 

The main design choices here are what divergences to use for the label-dependent term and the consistency term. Indeed, a difference between our loss and AugMix is the use of $\DJSnp$ for both terms. This is not an arbitrary choice, but has theoretical justifications when learning with noisy labels~\citep{englesson2021generalized}.

%% file: 4_results.tex
\section{Experiments}
\label{sec:results}
This section, empirically investigates the effectiveness of the proposed losses for learning with noisy labels, on synthetic~(Section~\ref{sec:synthetic-noise}) and real-world noise~(Section \ref{sec:real-noise}). In Section~\ref{sec:understanding}, we perform an ablation study to substantiate the importance of the consistency term, when going from $\DJSnp$ to $\DGJSnp$. All these additional experiments are done on the more challenging CIFAR-100 dataset. 

\textbf{Experimental Setup.} We use ResNet 34 and 50 for experiments on CIFAR and WebVision datasets respectively and optimize them using SGD with momentum.  
The complete details of the training setup can be found in Appendix \ref{sup:training-details}. Most importantly, we take three main measures to ensure a fair and reliable comparison throughout the experiments: 1) we reimplement all the loss functions we compare with in a single shared learning setup, 2) we use the same hyperparameter optimization budget and mechanism for all the prior works and ours, and 3) we train and evaluate five networks for individual results, where in each run the synthetic noise, network initialization, and data-order are differently randomized. The thorough  analysis is evident from the higher performance of CE in our setup compared to prior works. Where possible, we report mean and standard deviation and denote the statistically-significant top performers with student t-test.
\begin{figure}
    \centering
\begin{subfigure}[b]{0.23\textwidth} \centering 
    \includegraphics[width=1.0\textwidth]{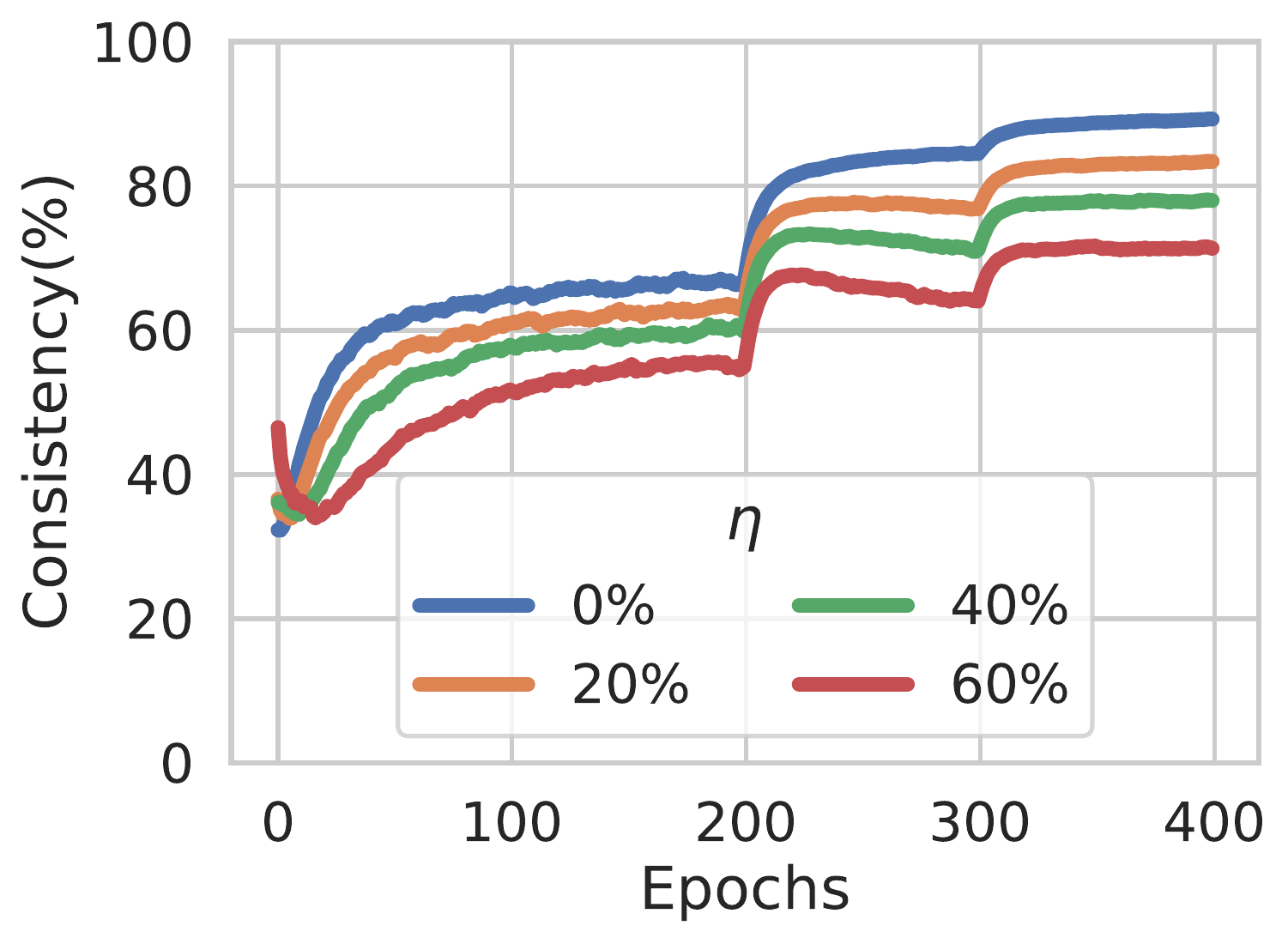} \caption{Consistency Clean} 
\end{subfigure} 
\hfill
\begin{subfigure}[b]{0.23\textwidth} \centering 
    \includegraphics[width=1.0\textwidth]{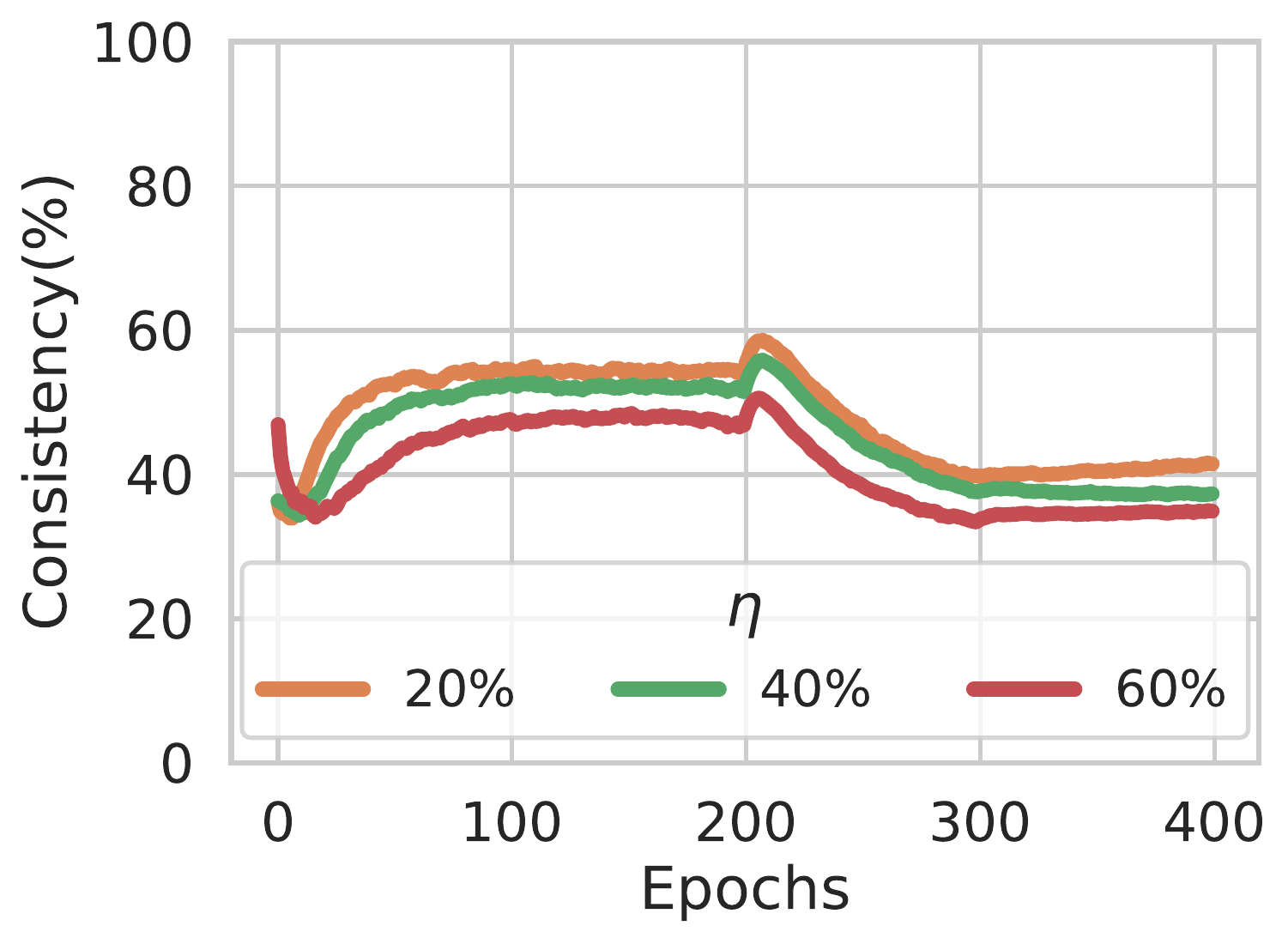} \caption{Consistency Noisy} 
\end{subfigure}
\caption{\textbf{Consistency of clean and noisy examples.} Consistency of correct~(clean) and mislabelled~(noisy) examples when overfitting to noise for various noise rates on CIFAR-100. The consistency is significantly degraded for the noisy data points. \label{fig:consistency-clean-noisy}\vspace{-0.2cm}}
\end{figure}
\subsection{Synthetic Noise Benchmarks: CIFAR}
\label{sec:synthetic-noise}
Here, we evaluate the proposed loss functions on the CIFAR datasets with two types of synthetic noise: symmetric and asymmetric, see Appendix \ref{sup:cifar-training-details} for details. 

We compare with other noise-robust loss functions such as 
Bootstrap~(BS)~\citep{Reed_arXiv_2014_bootstrapping}, Symmetric Cross-Entropy~(SCE)~\citep{Wang_ICCV_2019_Symmetric_CE}, and Generalized Cross-Entropy~(GCE)~\citep{Zhang_NeurIPS_2018_Generalized_CE}, 
We do not compare to methods that propose a full pipeline since, first, a conclusive comparison would require re-implementation and individual evaluation of several components and second, robust loss functions can be complementary to them.

\textbf{Results.}
Table \ref{tab:cifar} shows the results for
symmetric and asymmetric noise on CIFAR-10 and CIFAR-100. 
$\DGJSnp$ performs similarly or better than other methods for different noise rates, noise types, and data sets. Generally, $\DGJSnp$'s efficacy is more evident for the more challenging CIFAR-100 dataset. For example, on 60\% uniform noise on CIFAR-100, the difference between $\DGJSnp$ and the second best~(GCE) is 4.94 percentage points. 
Interestingly, the performance of $\DJSnp$ is consistently similar to the top performance of the prior works across different noise rates, types and datasets. 
Next, we test $\DGJSnp$ on a naturally-noisy dataset to see its efficacy in a real-world scenario.
\renewcommand{\arraystretch}{1.2}
\begin{table}[t!]
\tiny 
\caption{\label{tab:cifar} \textbf{Synthetic Noise Benchmark on CIFAR.} We \textit{reimplement} other noise-robust loss functions into the \textit{same learning setup} and ResNet-34, including Bootstrap~(BS), Generalized CE~(GCE), and Symmetric Cross Entropy~(SCE). We used \textit{same hyperparameter optimization budget and mechanism} for all the prior works and ours. Mean test accuracy and standard deviation are reported from five runs and the statistically-significant top performers are boldfaced. 
$\DGJSnp$ achieves state-of-the-art results for different noise rates, types, and datasets. Generally, $\DGJSnp$'s efficacy is more evident for the more challenging CIFAR-100 dataset.
}
\begin{center}
\tabcolsep=0.15cm 
\begin{tabular}{ @{}m{0.4cm} m{0.3cm} c c c c c@{}} 
 \toprule
 \multirow{2}{*}{Dataset} & \multirow{2}{*}{Loss} & No Noise & \multicolumn{2}{c}{Symmetric Noise Rate} & \multicolumn{2}{c}{Asymmetric Noise Rate} \\ \cmidrule(lr){3-3}\cmidrule(lr){4-5} \cmidrule(lr){6-7}
 & & 0\% & 20\% & 60\% & 20\% & 40\% \\
 \midrule
 \multirow{5}{5em}{C10} & CE & \textbf{95.77 $\pm$ 0.11} & 91.63 $\pm$ 0.27 & 81.99 $\pm$ 0.56 & 92.77 $\pm$ 0.24 & 87.12 $\pm$ 1.21 \\
 & BS & 94.58 $\pm$ 0.25 & 91.68 $\pm$ 0.32 & 82.65 $\pm$ 0.57 & 93.06 $\pm$ 0.25 & 88.87 $\pm$ 1.06 \\
 & SCE & \textbf{95.75 $\pm$ 0.16} & 94.29 $\pm$ 0.14 & 89.26 $\pm$ 0.37 &  93.48 $\pm$ 0.31 & 84.98 $\pm$ 0.76 \\
& GCE & \textbf{95.75 $\pm$ 0.14} & 94.24 $\pm$ 0.18 & 89.37 $\pm$ 0.27 & 92.83 $\pm$ 0.36 & 87.00 $\pm$ 0.99 \\
  & JS & \textbf{95.89 $\pm$ 0.10} & 94.52 $\pm$ 0.21 & 89.64 $\pm$ 0.15 & 92.18 $\pm$ 0.31 & 87.99 $\pm$ 0.55  \\
 & GJS & \textbf{95.91 $\pm$ 0.09} & \textbf{95.33 $\pm$ 0.18} & \textbf{91.64 $\pm$ 0.22} & \textbf{93.94 $\pm$ 0.25} & \textbf{89.65 $\pm$ 0.37} \\
\midrule
 \multirow{5}{5em}{C100} & CE & 77.60 $\pm$ 0.17 & 65.74 $\pm$ 0.22 & 44.42 $\pm$ 0.84 & 66.85 $\pm$ 0.32 & 49.45 $\pm$ 0.37  \\
 & BS & 77.65 $\pm$ 0.29 & 72.92 $\pm$ 0.50  & 53.80 $\pm$ 1.76 & 73.79 $\pm$ 0.43 & \textbf{64.67 $\pm$ 0.69} \\
 & SCE & 78.29 $\pm$ 0.24 & 74.21 $\pm$ 0.37  & 59.28 $\pm$ 0.58 & 70.86 $\pm$ 0.44 & 51.12 $\pm$ 0.37 \\
 & GCE & 77.65 $\pm$ 0.17 & 75.02 $\pm$ 0.24  & 65.21 $\pm$ 0.16 & 72.13 $\pm$ 0.39 & 51.50 $\pm$ 0.71 \\
 & JS & 77.95 $\pm$ 0.39 & 75.41 $\pm$ 0.28 & 64.36 $\pm$ 0.34 & 71.70 $\pm$ 0.36 & 49.36 $\pm$ 0.25 \\
 & GJS & \textbf{79.27 $\pm$ 0.29} & \textbf{78.05 $\pm$ 0.25}  & \textbf{70.15 $\pm$ 0.30} & \textbf{74.60 $\pm$ 0.47} & 63.70 $\pm$ 0.22 \\
 \bottomrule
\end{tabular}
\end{center}
\vspace{-0.3cm}
\end{table}
\subsection{Real-World Noise Benchmark: WebVision}
\label{sec:real-noise}
WebVision v1 is a large-scale image dataset collected by crawling Flickr and Google, which resulted in an estimated  20\% of noisy labels~\citep{li2017webvision}. There are 2.4 million images of the same thousand classes as ILSVRC12. Here, we use a smaller version called mini WebVision~\citep{Jiang18MentorNet} consisting of the first 50 classes of the Google subset.

\textbf{Results.}
Table \ref{tab:webvision}, as the common practice, reports the performances on the validation sets of WebVision and ILSVRC12~(first 50 classes). Both $\DJSnp$ and $\DGJSnp$ exhibits large margins with standard CE, especially using top-1 accuracy. Top-5 accuracy, due to its admissibility of wrong top predictions, can obscure the susceptibility to noise-fitting and thus indicates smaller but still significant improvements. 

The two state-of-the-art methods on this dataset were DivideMix\citep{Li_ICLR_2020_dividemix} and ELR+\citep{liu2020earlylearning}. Compared to our setup, both these methods use a stronger network~(Inception-ResNet-V2 vs ResNet-50), stronger augmentations~(Mixup vs color jittering) and co-train two networks instead of a single one. Furthermore, ELR+ uses an exponential moving average of weights and DivideMix treats clean and noisy labelled examples differently after separating them using Gaussian mixture models. Despite these differences, $\DGJSnp$ performs as good or better in terms of top-1 accuracy on WebVision and significantly outperforms ELR+ on ILSVRC12~(70.29 vs 74.33). 

So far, the experiments demonstrated the robustness of the proposed loss function via the significant improvements of final accuracy on noisy datasets. While this was central and informative, it is also important to investigate whether this improvement come from the properties that were argued for $\DGJSnp$. In what follows, we devise such experiments.
%
%
%
\begin{table}
\scriptsize
\tabcolsep=0.095cm 
\caption{\label{tab:webvision}\textbf{Real-world Noise Benchmark on WebVision.} Mean test accuracy and standard deviation from five runs are reported for the validation sets of (mini) WebVision and ILSVRC12. 
Results marked with $\dagger$ are from \citep{zheltonozhskii2021contrast}. DivideMix with $\star$ uses a ResNet-50~(used by CE, $\DJSnp$, $\DGJSnp$) instead of InceptionResNet-V2. \vspace{0.2cm}}
\centering
\begin{tabular}{lcccc}\toprule
\multirow{2}{4em}{Method}  & \multicolumn{2}{c}{WebVision} & \multicolumn{2}{c}{ILSVRC12}
\\\cmidrule(lr){2-3}\cmidrule(lr){4-5}
        & Top 1  & Top 5     & Top 1  & Top 5 \\\midrule
ELR+{$\dagger$} &  $\bm{77.78}$ & $\bm{91.68}$ & 70.29 & 89.76 \\
DivideMix{$\dagger$} & 77.32 & $\bm{91.64}$ & $\bm{75.20}$  & 90.84  \\
DivideMix{$\star\dagger$} & 76.32 $\pm$ 0.36 & 90.65 $\pm$ 0.16 & 74.42 $\pm$ 0.29 & $\bm{91.21 \pm 0.12}$ \\
 \midrule
CE & 70.69 $\pm$ 0.66 & 88.64 $\pm$ 0.17 & 67.32 $\pm$ 0.57 & 88.00 $\pm$ 0.49 \\
 JS & 74.56 $\pm$ 0.32 & 91.09 $\pm$ 0.08 & 70.36 $\pm$ 0.12 & 90.60 $\pm$ 0.09  \\
GJS & $\bm{77.99 \pm 0.35}$ & 90.62 $\pm$ 0.28 & 74.33 $\pm$ 0.46 & 90.33 $\pm$ 0.20 \\
\bottomrule
\end{tabular}
\vspace{-0.3cm}
\end{table}
%
\vspace{-0.1cm}
\subsection{Towards a Better Understanding of GJS}
\label{sec:understanding}
\textbf{Is the improvements of GJS over JS due to mean prediction or consistency?}
The difference between $\DJSnp$ and $\DGJSnp$ is the mean prediction in the label-dependent term and the consistency term. In Table~\ref{tab:gjs-wo-consistency}, we study these differences by using $\DGJSnp$ without the consistency term, i.e., $\DJSnp(\vy,\vm)$.
The results suggest that the improvement of $\DGJSnp$ over $\DJSnp$ can be crucially attributed to the consistency term. 

\textbf{Is GJS mostly helping the clean or noisy examples?} 
To better understand the improvements of $\DGJSnp$ over $\DJSnp$, we perform an ablation with different losses for clean and noisy examples, see Table~\ref{tab:cifar-concleannoisy}. 
Using $\DGJSnp$ instead of $\DJSnp$ improves performance in all cases. Importantly, using $\DGJSnp$ only for the noisy examples performs significantly better than only using it for the clean examples~(74.1 vs 72.9). The best result is achieved when using $\DGJSnp$ for both clean and noisy examples but still close to the noisy-only case~(74.7 vs 74.1).

\textbf{Additional Experiments.} In Appendix \ref{sup:sec:additional-experiments}, we study the role of augmentations and the training behavior of $\DGJSnp$.

\begin{table}
    \centering
    \captionof{table}{\textbf{Effect of Consistency.} Validation accuracy for $\DJSnp$, $\DGJSnp$ w/o the consistency term in Equation~\ref{eq:gjs-loss}, and $\DGJSnp$ for $40\%$ noise on the CIFAR-100 dataset. Using the mean of two predictions in the $\DJSnp$ loss does not improve performance. On the other hand, adding the consistency term significantly helps.\vspace{0.2cm}
    \label{tab:gjs-wo-consistency}}
    \scriptsize
    \begin{tabular}{ l c }  \toprule
        Method & Accuracy \\ \midrule
        $\DJS(\vy,\vp_{1})$ & 71.0\\ 
        $\DJS(\vy,\vm)$ & 68.7  \\ 
        $\DGJSnp$ & 74.3 \\ 
        \bottomrule 
    \end{tabular}
    \vspace{-0.2cm}
\end{table}
\begin{table}
\vspace{-0.3cm}
        \centering
        \scriptsize
        \tabcolsep=0.16cm 
        \captionof{table}{\label{tab:cifar-concleannoisy} \textbf{Effect of }$\bm{\DGJSnp}$\textbf{.} Validation accuracy when using different loss functions for clean and noisy examples of the CIFAR-100 training set with 40\% symmetric noise. Noisy examples benefit significantly more from $\DGJSnp$ than clean examples~(74.1 vs 72.9).
        \vspace{0.2cm}}
        \begin{tabular}{c c c c c }\toprule
        \multicolumn{2}{c}{Method} & \multicolumn{3}{c}{$\pi$} \\ \cmidrule(lr){1-2}\cmidrule(lr){3-5}
        Clean & Noisy & 0.1 & 0.5 & 0.9
        \\\midrule
        JS & JS & 70.0 & $\bm{71.5}$ & 55.3 \\
        GJS & JS & 72.6 & $\bm{72.9}$ & 70.2 \\
        JS & GJS & 71.0 & $\bm{74.1}$ & 68.0 \\
        GJS & GJS & 71.3 & $\bm{74.7}$ & 73.8 \\ \bottomrule
        \end{tabular}
    \vspace{-0.3cm}
\end{table}

%% file: 2_related_works_short.tex
\section{Related Works}
\vspace{-0.1cm}
\label{sec:rel_works}
\textit{Consistency regularization} is a recent technique that imposes smoothness in the learnt function for semi-supervised learning~\citep{Oliver_arXiv_2018_Realistic_Eval_SSL} and recently for noisy data~\citep{Li_ICLR_2020_dividemix}. These works use complex pipelines for such regularization. $\DGJSnp$ encourages consistency in a simple way that exhibits other desirable properties for learning with noisy labels. 
Importantly, \citet{hendrycks2020augmix} used Jensen-Shannon-based loss functions to improve test-time robustness to image corruptions which further verifies the general usefulness of $\DGJSnp$. In this work, we study such loss functions for the different goal of \textit{training-time label-noise robustness}. In this context, our thorough analytical and empirical results are, to the best of our knowledge, novel.

Recently, loss functions with \textit{information-theoretic} motivations have been proposed~\citep{Xu_NeurIPS_2019_Information_Theoretic_Mutual_Info_Loss,Wei_ICLR_2021_f_Divergence}. $\DJSnp$, with apparent information-theoretic interpretation, has a strong connection to those. The latter is a close concurrent work but takes a different and complementary angle. The connection of these works can be a fruitful future direction.

%% file: 5_future_work_limitations.tex
\vspace{-0.25cm}
\section{Final Remarks}
\vspace{-0.1cm}
\label{sec:final-remarks}
Overall, we believe the paper provides useful and novel observations and informative empirical evidence for the usefulness of a JS divergence-based consistency loss for learning under noisy data that achieve state-of-the-art results. At the same time it opens interesting future directions.
\vspace{-0.3cm}

%% file: acknowledgement.tex
\paragraph{Acknowledgement.} This work was partially supported by the Wallenberg AI, Autonomous Systems and Software Program (WASP) funded
by the Knut and Alice Wallenberg Foundation.

%% file: 7_appendix.tex
\appendix
\onecolumn

\section{Training Details}
\label{sup:training-details}
Our method and the baselines use the same training settings, which are described in detail here.
\subsection{CIFAR}
\label{sup:cifar-training-details}
\textbf{General Training Details.} For all the results on the CIFAR datasets, we use a PreActResNet-34 with a standard SGD optimizer with Nesterov momentum, and a batch size of 128. For the network, we use three stacks of five residual blocks with 32, 64, and 128 filters for the layers in these stacks, respectively. The learning rate is reduced by a factor of 10 at 50\% and 75\% of the total 400 epochs. For data augmentation, we use RandAugment~\citep{cubuk2019randaugment} with $N=1$ and $M=3$ using random cropping~(size 32 with 4 pixels as padding), random horizontal flipping, normalization and lastly Cutout~\citep{devries2017cutout} with length 16. We set random seeds for all methods to have the same network weight initialization, order of data for the data loader, train-validation split, and noisy labels in the training set. We use a clean validation set corresponding to 10\% of the training data. A clean validation set is commonly provided with real-world noisy datasets~\citep{li2017webvision,li2019learning}. Any potential gain from using a clean instead of a noisy validation set is the same for all methods since all share the same setup. 

\textbf{Noise Types.} For symmetric noise, the labels are, with probability $\eta$, re-sampled from a uniform distribution over all labels. For asymmetric noise, we follow the standard setup of \citet{patrini2017making}. For CIFAR-10, the labels are modified, with probability $\eta$, as follows: \textit{truck}~$\to$~\textit{automobile}, \textit{bird}~$\to$~\textit{airplane}, \textit{cat}~$\leftrightarrow$~\textit{dog}, and \textit{deer}~$\to$~\textit{horse}. For CIFAR-100, labels are, with probability $\eta$, cycled to the next sub-class of the same ``super-class'', \eg the labels of super-class ``vehicles 1'' are modified as follows: \textit{bicycle}~$\rightarrow$~$\textit{bus}$~$\rightarrow$~\textit{motorcycle}~$\rightarrow$~\textit{pickup truck}~$\rightarrow$~\textit{train}~$\rightarrow$~\textit{bicycle}.

\textbf{Search for learning rate and weight decay.} We do a separate hyperparameter search for learning rate and weight decay on 40\% noise using both asymmetric and symmetric noises on CIFAR datasets. For CIFAR-10, we search for learning rates in $[0.001, 0.005, 0.01, 0.05, 0.1]$ and weight decays in $[1e-4, 5e-4, 1e-3]$. The method-specific hyperparameters used for this search were 0.9, 0.7, (0.1,1.0), 0.7, (1.0,1.0), 0.5, 0.5 for BS($\beta$), LS($\epsilon$), SCE($\alpha,\beta$), GCE($q$), NCE+RCE($\alpha,\beta$), JS($\pi$) and GJS($\pi$), respectively. For CIFAR-100, we search for learning rates in $[0.01, 0.05, 0.1, 0.2, 0.4]$ and weight decays in $[1e-5, 5e-5, 1e-4]$. The method-specific hyperparameters used for this search were 0.9, 0.7, (6.0,0.1), 0.7, (10.0,0.1), 0.5, 0.5 for BS($\beta$), LS($\epsilon$), SCE($\alpha,\beta$), GCE($q$), NCE+RCE($\alpha,\beta$), JS($\pi$) and GJS($\pi$), respectively. Note that, these fixed method-specific hyperparameters for both CIFAR-10 and CIFAR-100 are taken from their corresponding papers for this initial search of learning rate and weight decay but they will be further optimized systematically in the next steps.

\textbf{Search for method-specific parameters.} We fix the obtained best learning rate and weight decay for all other noise rates, but then for each noise rate/type, we search for method-specific parameters. For the methods with a single hyperparameter, BS~($\beta$), LS~($\epsilon$), GCE~($q$), JS~($\pi$), GJS~($\pi$), we try values in $[0.1, 0.3, 0.5, 0.7, 0.9]$. On the other hand, NCE+RCE and SCE have three hyperparameters, \ie $\alpha$ and $\beta$ that scale the two loss terms, and $A\coloneqq \log(0)$ for the RCE term. We set $A=\log{(1e-4)}$ and do a grid search for three values of $\alpha$ and two of beta $\beta$~(six in total) around the best reported parameters from each paper.\footnote{We also tried using $\beta=1-\alpha$, and mapping the best parameters from the papers to this range, combined with a similar search as for the single parameter methods, but this resulted in worse performance.}
 
\textbf{Test evaluation.} The best parameters are then used to train on the full training set with five different seeds. The final parameters that were used to get the results in Table \ref{tab:cifar} are shown in Table \ref{tab:cifar-hps}. 

 \begin{table*}[t!]
 \scriptsize
 \caption{\label{tab:cifar-hps} \textbf{Hyperparameters for CIFAR.} A hyperparameter search over learning rates and weight decays, was done for 40\% noise on both symmetric and asymmetric noise for the CIFAR datasets. The best parameters for each method are shown in this table, where the format is [learning rate, weight decay]. The hyperparameters for zero percent noise uses the same settings as for the symmetric noise. For the best learning rate and weight decay, another search is done for method-specific hyperparameters, and the best values are shown here. For methods with a single hyperparameter, the value correspond to their respective hyperparameter, i.e., BS~($\beta$), GCE~($q$), JS~($\pi$), GJS~($\pi$). For SCE the value correspond to [$\alpha$, $\beta$].
 }
 \begin{center}
 \tabcolsep=0.12cm 
 \input{Figures/app_table_hp_all}
 \end{center}
 \end{table*}

\subsection{WebVision}
All methods train a randomly initialized ResNet-50 model from PyTorch using the SGD optimizer with Nesterov momentum, and a batch size of 32 for $\DGJSnp$ and 64 for CE and $\DJSnp$. For data augmentation, we do a random resize crop of size 224, random horizontal flips, and color jitter~(torchvision ColorJitter transform with brightness=0.4, contrast=0.4, saturation=0.4, hue=0.2). We use a fixed weight decay of $1e-4$ and do a grid search for the best learning rate in $[0.1, 0.2, 0.4]$ and $\pi \in [0.1, 0.3, 0.5, 0.7, 0.9]$. The learning rate is reduced by a multiplicative factor of $0.97$ every epoch, and we train for a total of 300 epochs. 
The best starting learning rates were 0.4, 0.2, 0.1 for CE, JS and GJS, respectively. Both $\DJSnp$ and $\DGJSnp$ used $\pi=0.1$. With the best learning rate and $\pi$, we ran four more runs with new seeds for the network initialization and data loader.

\section{Loss Implementation Details}
\textbf{Loss.} We implement the Jensen-Shannon divergence using the definitions based on KL divergence, see Equation \ref{eq:GJSasKL}. To make sure the gradients are propagated through the target argument, we do not use the built-in KL divergence in PyTorch. Instead, we write our own based on the official implementation. 

\textbf{Scaling.} The value of the $\DJS$ divergence becomes small as the hyperparameter $\pi$ approaches 0 and 1. To counteract this, we divide both $\DJSnp$ and $\DGJSnp$ losses by a constant factor, $-(1-\pi)\log(1-\pi)$. Clearly, this scaling is equivalent to a scaling of the learning rate.

\section{Consistency Measure}
\label{sup:sec:consistencyMeasure}
 In this section, we provide more details about the consistency measure used in \eg Figure \ref{fig:consistency-observation}. To be independent of any particular loss function, we considered a measure similar to standard Top-1 accuracy. We measure the ratio of samples that predict the same class on both the original image and an augmented version of it
 \begin{align}
     \frac{1}{N}\sum_{i=1}^N \mathbbm{1}\big(\argmax_y f(\vx_i) = \argmax_y f(\tilde{\vx}_i)\big)
 \end{align}
 where the sum is over all the training examples, and $\mathbbm{1}$ is the indicator function, the argmax is over the predicted probability of $K$ classes, and $\tilde{\vx}_i$ is an augmented version of $\vx_i$. Notably, this measure does not depend on the labels. 
 
 In the experiments in Figure~\ref{fig:consistency-observation} and \ref{fig:consistency-clean-noisy}, the original images are only normalized, while the augmented images use the same augmentation strategy as the benchmark experiments in Section \ref{sup:cifar-training-details}.

\section{Additional Experiments}
\label{sup:sec:additional-experiments}
\subsection{Augmentations}
As described in Section \ref{sup:cifar-training-details}, our augmentation strategy is composed of several transformations: random crop, horizontal flips, CutOut, and RandAugment. Here, we study the noise-robustness of our method when removing some of these transformations. We either remove CutOut, RandAugment, or both~(which we denote by ``weak''). See Table~\ref{tab:cifar-augs} for the results of 40\% symmetric and asymmetric noise rates on CIFAR-100. While more transformations help improve robustness for all methods, it is not required for $\DGJSnp$ to perform well.

\begin{table*}[t]
        \centering
        \scriptsize
        \tabcolsep=0.16cm 
        \captionof{table}{\label{tab:cifar-augs} \textbf{Effect of Augmentation Strategy.} Validation accuracy for training w/o CutOut(-CO) or w/o RandAug(-RA) or w/o both(weak) on 40\% symmetric and asymmetric noise on CIFAR-100. All methods improves by stronger augmentations. GJS performs best for all types of augmentations.}
        \begin{tabular}{p{0.9cm} >{\centering}p{0.365cm} >{\centering}p{0.45cm} >{\centering}p{0.45cm} >{\centering}p{0.365cm} >{\centering}p{0.365cm} >{\centering}p{0.45cm} >{\centering}p{0.45cm} >{\centering\arraybackslash}p{0.365cm}}\toprule
        \multirow{2}{4em}{Method} & \multicolumn{4}{c}{Symmetric} & \multicolumn{4}{c}{Asymmetric} \\
        \cmidrule(lr){2-5} \cmidrule(lr){6-9}
        & Full & -CO & -RA & Weak & Full & -CO & -RA & Weak
        \\\midrule
        GCE & 70.8 & 64.2 & 64.1 & 58.0 & 51.7 & 44.9 & 46.6 & 42.9  \\
        NCE+RCE & 68.5 & 66.6 & 68.3 & 61.7 & 57.5 & 52.1 & 49.5 & 44.4  \\
        GJS & $\bm{74.8}$ & $\bm{71.3}$ & $\bm{70.6}$ & $\bm{66.5}$ & \textbf{62.6} & \textbf{56.8} & \textbf{52.2} & \textbf{44.9}  \\ \bottomrule
        \end{tabular}
\end{table*}

\subsection{Training Behavior of Networks using GJS}
\begin{figure}[h]
    \captionsetup[subfigure]{aboveskip=-1pt,belowskip=-1pt}
    \centering
     \subcaptionbox[b]{CE - Clean}
    {\includegraphics[width=0.23\textwidth]{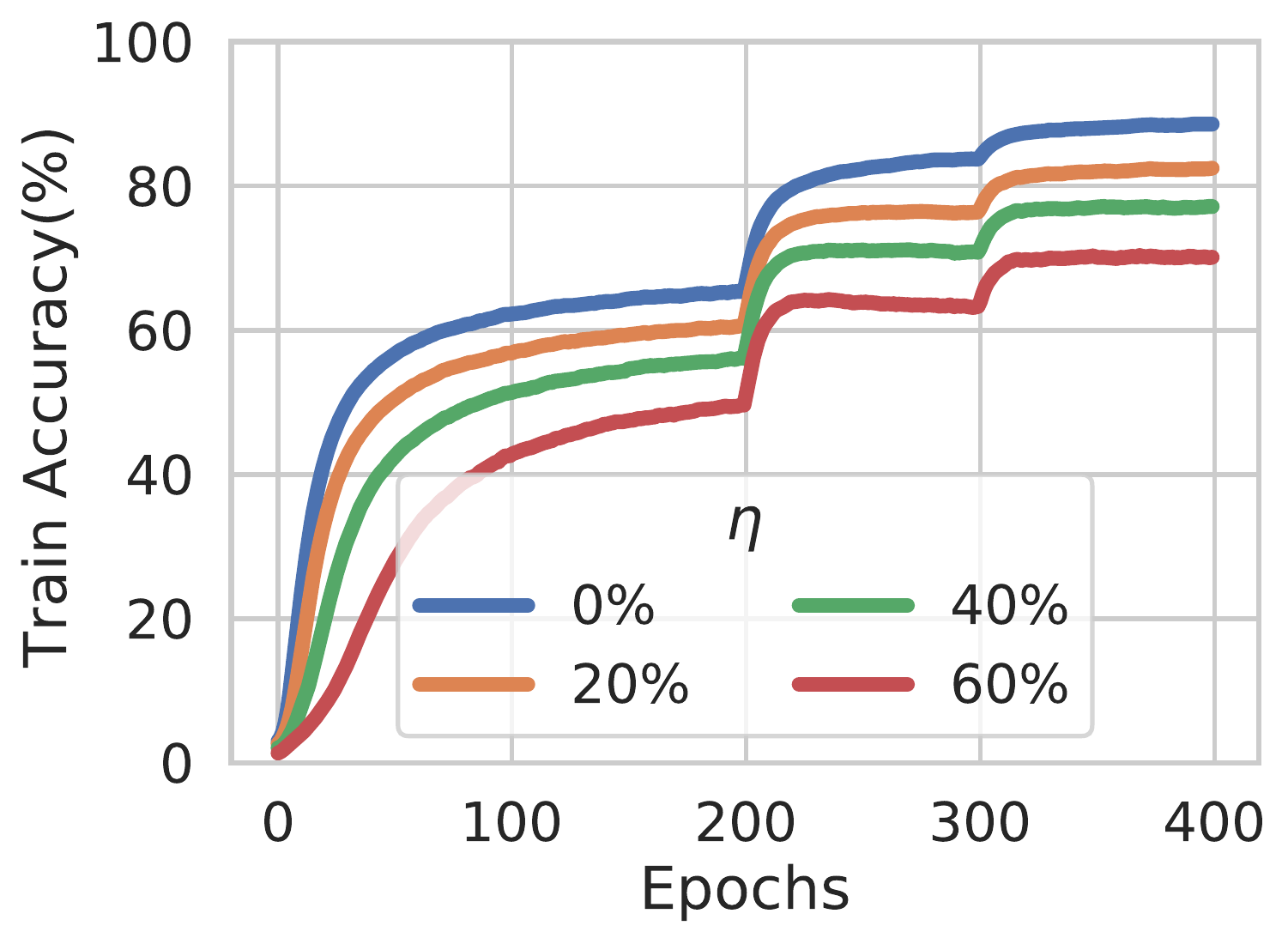}}
      \subcaptionbox[b]{CE - Noisy}
    {\includegraphics[width=0.23\textwidth]{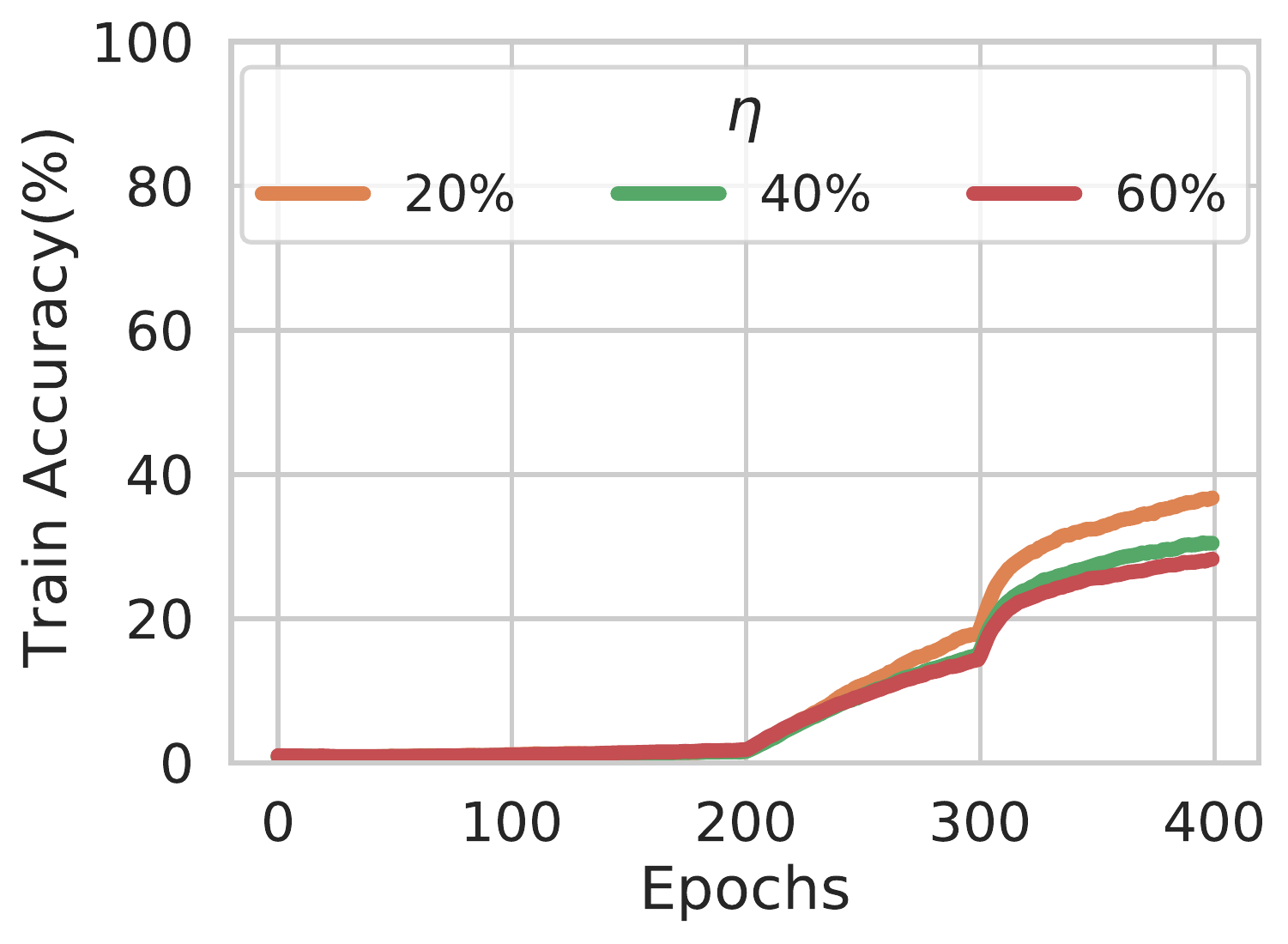}}
   \subcaptionbox[b]{CE - Clean}
    {\includegraphics[width=0.23\textwidth]{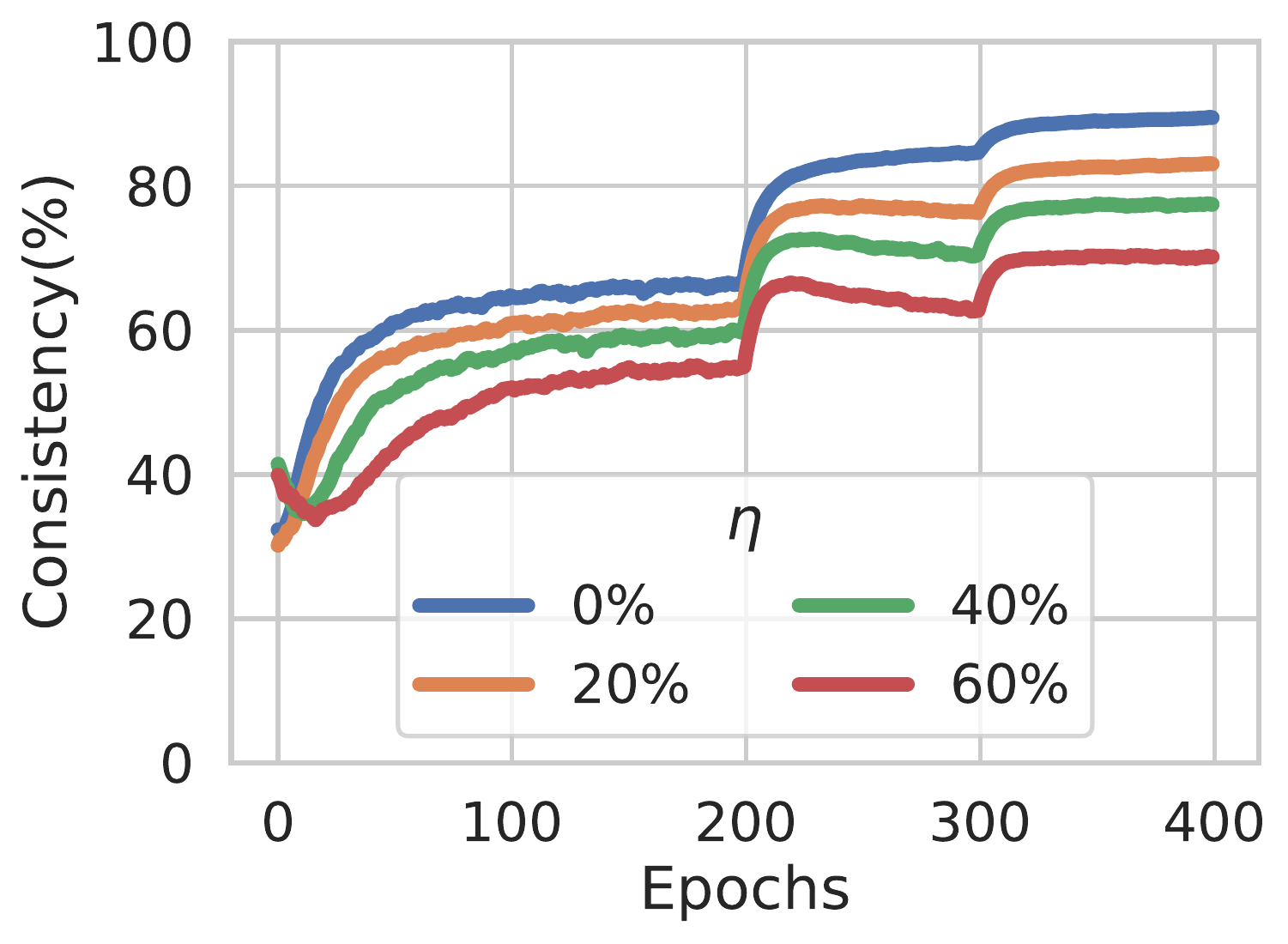}}
    \subcaptionbox[b]{CE - Noisy}
      {\includegraphics[width=0.23\textwidth]{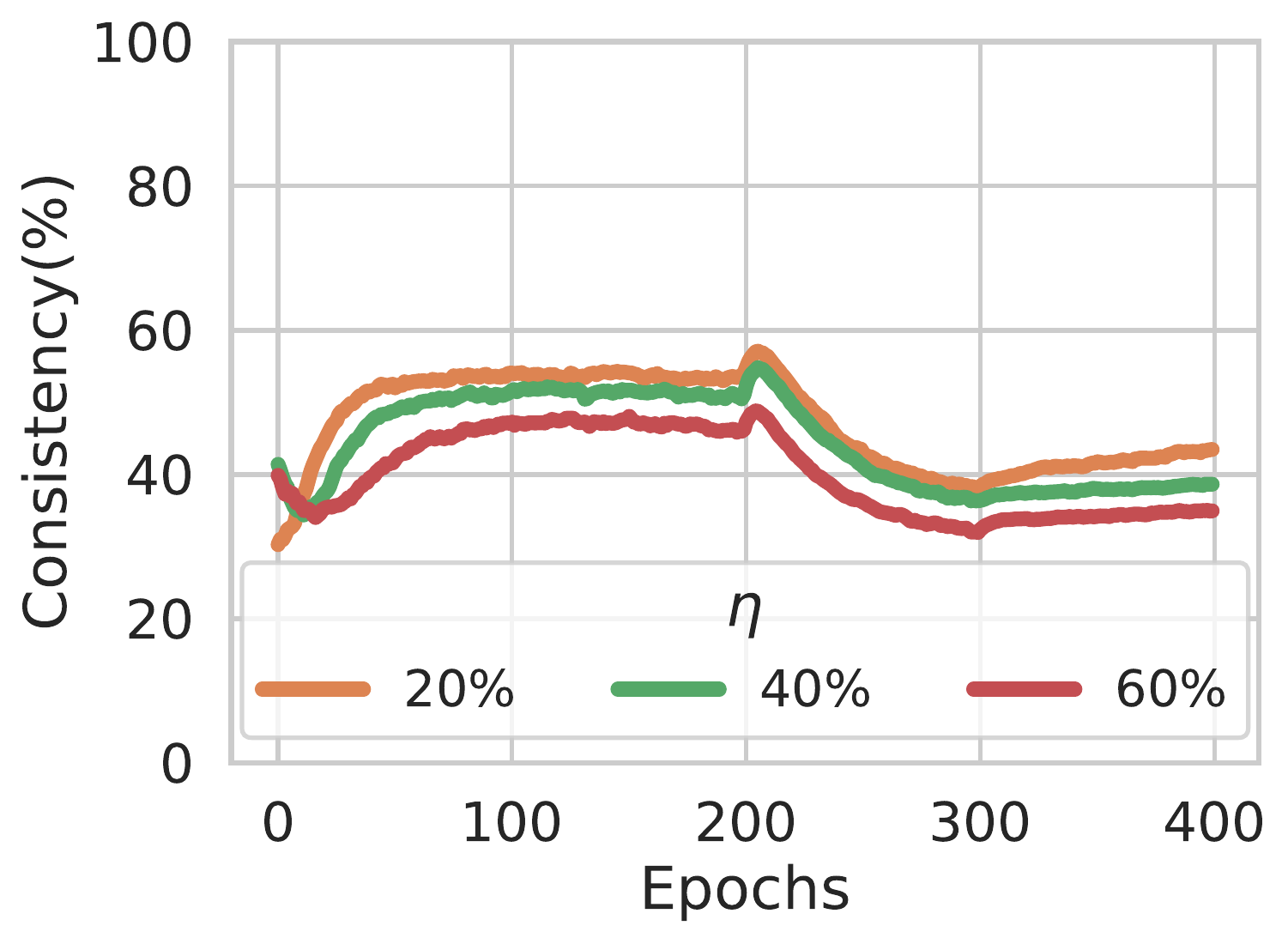}}
    \label{fig:piAblation}
     \subcaptionbox[b]{GJS - Clean}
    {\includegraphics[width=0.23\textwidth]{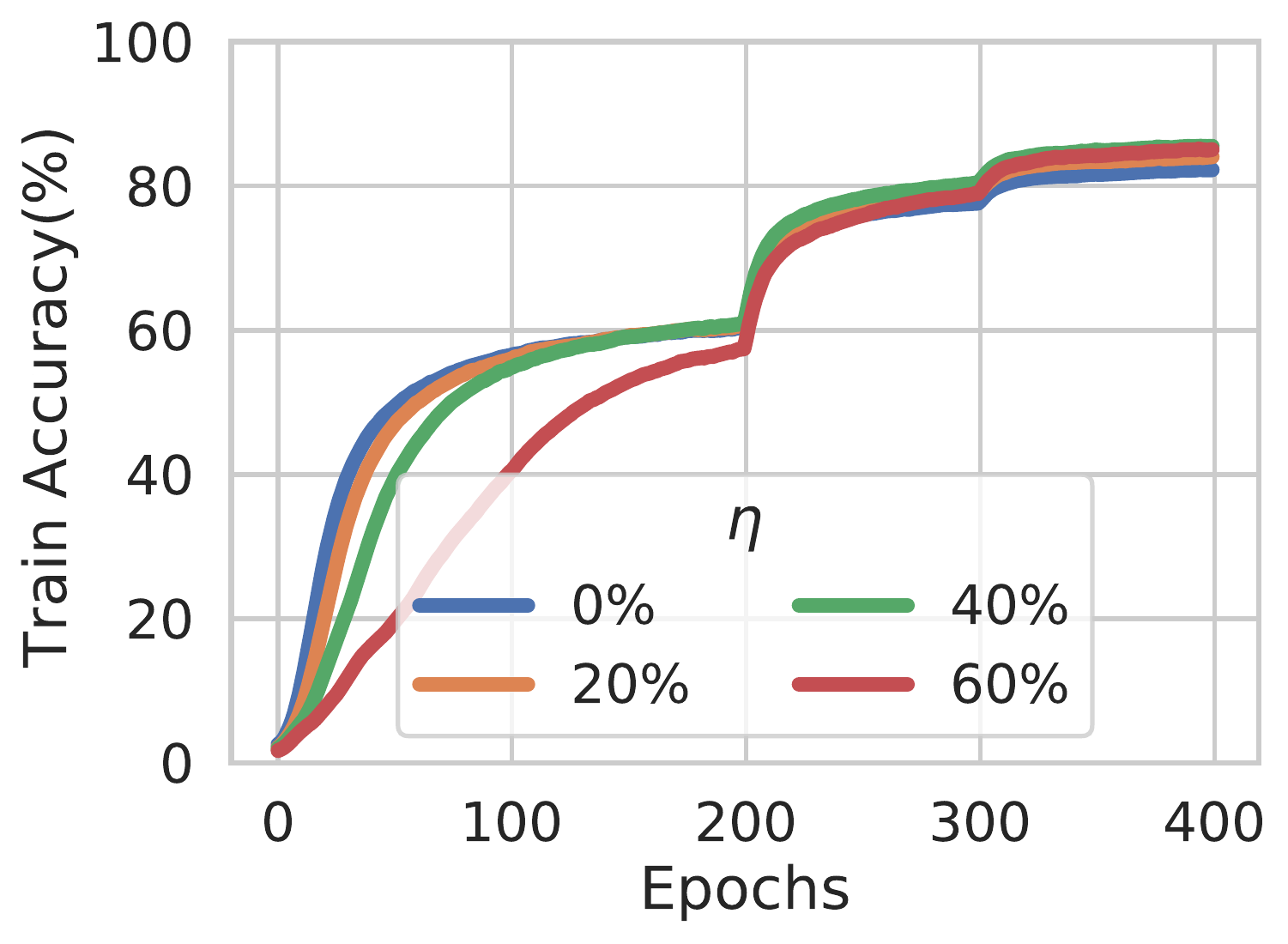}}
      \subcaptionbox[b]{GJS - Noisy}
    {\includegraphics[width=0.23\textwidth]{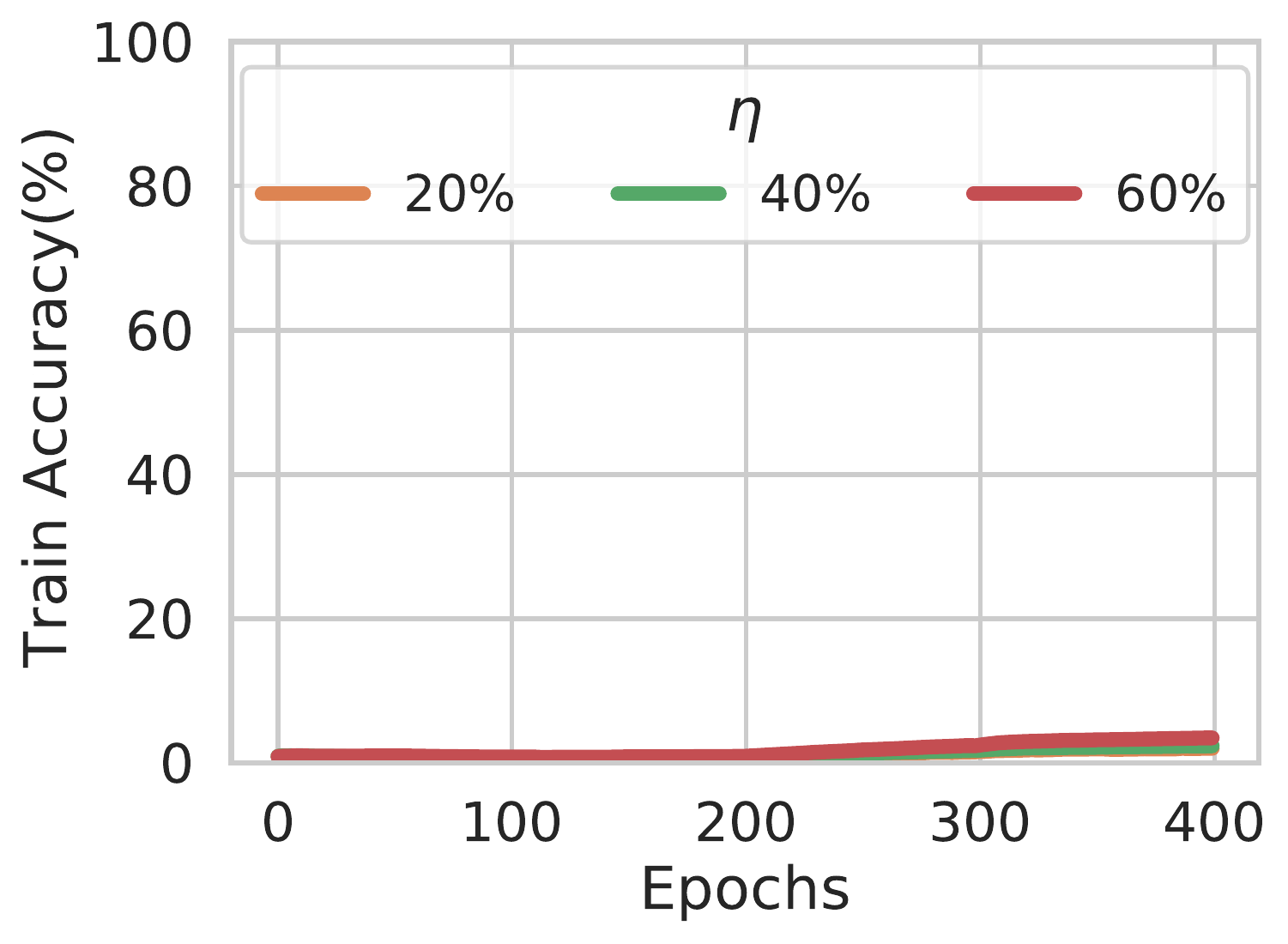}}
   \subcaptionbox[b]{GJS - Clean}
    {\includegraphics[width=0.23\textwidth]{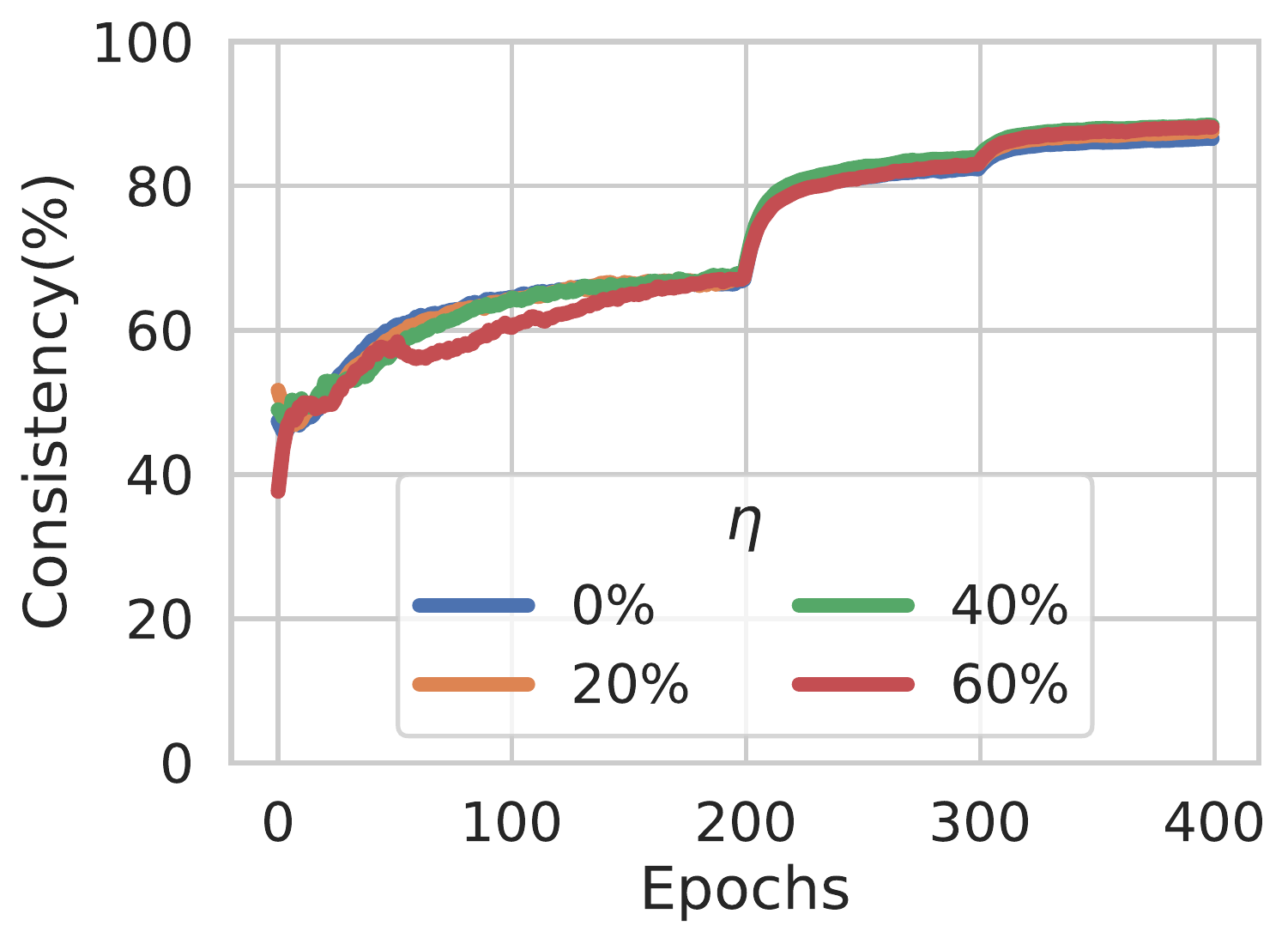}}
    \subcaptionbox[b]{GJS - Noisy}
      {\includegraphics[width=0.23\textwidth]{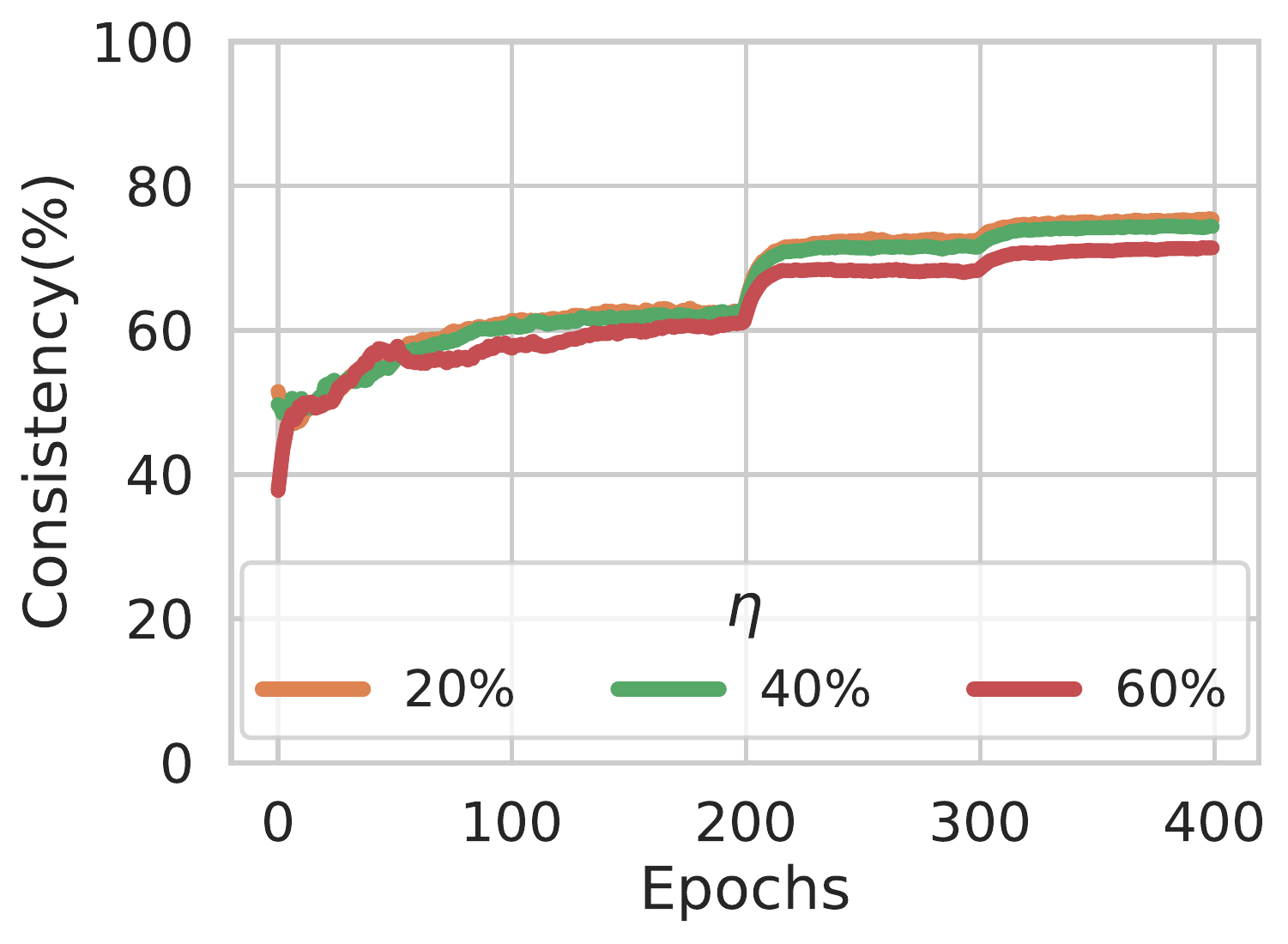}}
            \caption{\textbf{Evolution of training accuracy and consistency for clean and noisy examples.} Consistency and training accuracy of correct~(clean) and mislabelled~(noisy) examples when overfitting to noise for various symmetric noise rates on CIFAR-100 for the CE and $\DGJSnp$ losses. $\DGJSnp$ uses $\pi=0.9$, and both CE and GJS use the same learning rate~($0.2$) and weight decay~($5e-5$) for all noise rates. Using $\DGJSnp$ as a loss keeps the network's predictions consistent~(\eg (d) and (h)) and significantly reduces the overfitting to noise~((b)~and~(f)). Interestingly, $\DGJSnp$ also significantly improves the clean training accuracy when learning with noisy labels~((a) and (e)). \label{fig:train-acc-consistency} \vspace{-0.7cm}
        }
\end{figure}

In Section \ref{sec:consistency-obs}, we observed that networks become less consistent when trained with the CE loss on noisy data, especially the consistency of predictions on the noisy labelled examples. To better understand the improvements in robustness that $\DGJSnp$ brings to the learning dynamics of deep networks, we compare CE and $\DGJSnp$ in Figure \ref{fig:train-acc-consistency} in terms of \textit{training accuracy} and consistency for clean and noisy labelled examples. For clean~(correctly) labelled examples, $\DGJSnp$ improves the accuracy and consistency to be as good as without any noise. For noisy~(mislabelled) examples, we observe that $\DGJSnp$ significantly reduces the overfitting to noisy labels and keeps improving the consistency.

%% file: Figures/app_table_hp_all.tex
\begin{tabular}{ @{}l l c c c c c c c c c@{}} 
 \toprule
 \multirow{3}{4em}{Dataset} & \multirow{3}{4em}{Method} & \multicolumn{2}{c}{Learning Rate \& Weight Decay} & \multicolumn{7}{c}{Method-specific Hyperparameters}
 \\
  \cmidrule(lr){3-4} \cmidrule(lr){5-9}
 & & \multicolumn{1}{c}{Sym Noise} & \multicolumn{1}{c}{Asym Noise} & \multicolumn{1}{c}{No Noise} & \multicolumn{2}{c}{Sym Noise} & \multicolumn{2}{c}{Asym Noise} \\ 
 \cmidrule(lr){3-3} \cmidrule(lr){4-4} \cmidrule(lr){5-5} \cmidrule(lr){6-9} \cmidrule(lr){10-11}
 & & 20-60\% & 20-40\% & 0\% & 20\% & 60\% & 20\% & 40\%   \\
 \midrule
 \multirow{8}{5em}{CIFAR-10} & CE & [0.05, 1e-3] & [0.1, 1e-3] & - & - & - & - & -\\
 & BS & [0.1, 1e-3] & [0.1, 1e-3] & 0.5 & 0.5 & 0.7 & 0.7 & 0.5 \\
 & SCE & [0.01, 5e-4] & [0.05, 1e-3] & [0.2, 0.1] & [0.05, 0.1] & [0.2, 1.0] & [0.1, 0.1] & [0.2, 1.0] \\
 & GCE & [0.01, 5e-4] & [0.1, 1e-3] & 0.5 & 0.7 & 0.7 & 0.1 & 0.1 \\
 & JS & [0.01, 5e-4] & [0.1, 1e-3] & 0.1 & 0.7 & 0.9 & 0.3 & 0.3 \\
 & GJS & [0.1, 5e-4] & [0.1, 1e-3] & 0.5 & 0.3 & 0.1 & 0.3 & 0.3 \\
\midrule
 \multirow{8}{5em}{CIFAR-100} & CE & [0.4, 1e-4] & [0.2, 1e-4] & - & - & - & - & - \\
 & BS & [0.4, 1e-4] & [0.4, 1e-4] & 0.7 & 0.5 & 0.5 & 0.3 & 0.3  \\
 & SCE & [0.2, 1e-4] & [0.4, 5e-5] & [0.1, 0.1] & [0.1, 0.1] & [0.1, 1.0] & [0.1, 1.0] & [0.1, 1.0] \\
 & GCE & [0.4, 1e-5] & [0.2, 1e-4] & 0.5 & 0.5 & 0.7 & 0.7 & 0.7 \\
 & JS & [0.2, 1e-4] & [0.1, 1e-4] & 0.1 & 0.1 & 0.5 & 0.5 & 0.5 \\
 & GJS & [0.2, 5e-5] & [0.4, 1e-4] & 0.3 & 0.3 & 0.9 & 0.5 & 0.1 \\
 \bottomrule
\end{tabular}